\documentclass[a4paper, 10pt, conference]{cssconf}
\IEEEoverridecommandlockouts
\usepackage{hyperref} 
\usepackage{url}

\usepackage{epsfig} % for postscript graphics files
\usepackage{amsmath} % assumes amsmath package installed

\title{\LARGE \bf Optimizing wearable assistive devices with neuromuscular models and optimal control}

\author{Manish Sreenivasa, Matthew Millard, Paul Manns and Katja Mombaur
\thanks{This work was supported by the European Union's Horizon 2020 research and innovation programme under grant agreement No. 687662 (SPEXOR project).}
\thanks{Authors are with the Optimization in Robotics \& Biomechanics group, Interdisciplinary Center for Scientific Computing, Heidelberg University, Berlinerstr. 45, 69120 Heidelberg, Germany (e-mail: {\tt\small manish.sreenivasa@iwr.uni-heidelberg.de}).}
\thanks{This is the author's version of the abstract that has been published on SpringerLink. The final publication is available on: \url{https://link.springer.com/chapter/10.1007/978-3-319-46669-9_103}}
}

\begin{document}
\maketitle
\thispagestyle{empty}
\pagestyle{empty}

%%%%%%%%%%%%%%%%%%%%%%%%%%%%%%%%%%%%%%%%%%%%
\begin{abstract}
The coupling of human movement dynamics with the function and design of wearable assistive devices is vital to better understand the interaction between the two. Advanced neuromuscular models and optimal control formulations provide the possibility to study and improve this interaction. In addition, optimal control can also be used to generate predictive simulations that generate novel movements for the human model under varying optimization criterion. 
\end{abstract}
\section{Introduction}
In this paper we provide an overview of the methods involved in the modeling of the human-exoskeleton system, as well as the solution process to determine optimal neural inputs and exoskeleton design parameters. 
\section{Materials and Methods}
\label{sec:dynamicHuman}
\begin{figure}[h]
\centering
\includegraphics[width=0.85\linewidth]{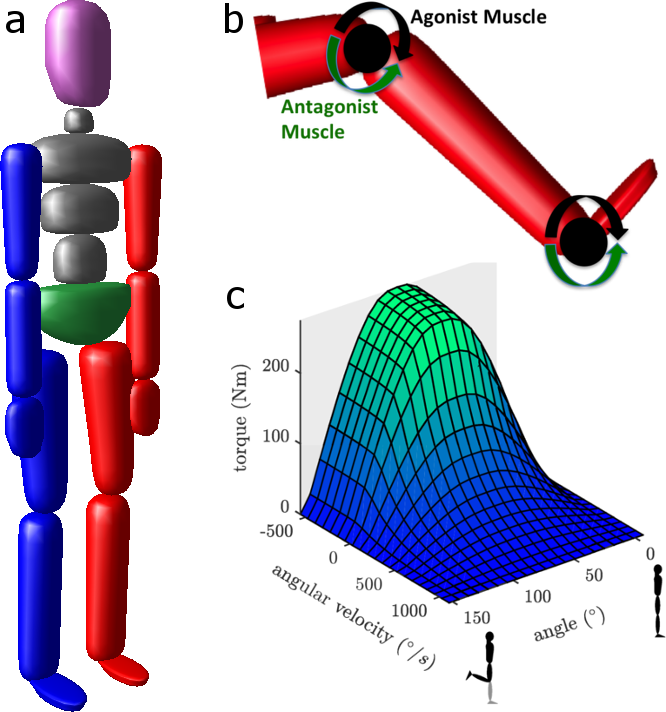}
 \caption{(a) 3D Human model (b) Illustration of agonist-antagonist torque muscles that actuate each rotational DoF (c) Smooth approximation of torque muscle dynamics using 2D $5^{th}$ order Bezier curves.}
\label{fig:humanModel}
\end{figure}

We model the human body as an articulated multi-body system, with each joint having 1 to 3 rotational degrees of freedom (DoFs). The model shown in Fig. \ref{fig:humanModel}a consists of a total of 33 rotational DoFs with an additional 6 DoFs for the non-actuated root joint. Each rotational DoF is actuated by a pair of agonist-antagonist ``torque muscles" \cite{Anderson2007},  which represent the combined torques being generated by muscle forces in that direction, Fig. \ref{fig:humanModel}b.
\subsection{Subject-specificity}
\label{subsec:subjectSpecific}
Depending on the application, the kinematic and dynamic parameters of the human model may be further specified towards individual characteristics. For example, for healthy participants it may be sufficient to adjust joint centers using graphical interfaces such as PUPPETEER \cite{Felis2016}, and segment masses and inertia from regression equations \cite{DeLeva1996}. For pathological cases, it may be necessary to develop patient specific models from functional analysis of motion capture data \cite{Gamage2002} as well as segmentation from MRI images \cite{Sreenivasa2016}.
\subsection{Neuromuscular Dynamics}
\label{subsec:neuromuscular}
The concept of agonist-antagonist torque muscles \cite{Anderson2007} along with excitation-activation dynamics \cite{Thelen2003} provides a good balance between model complexity and the ability to simulate physiologically realistic (aggregated) muscle and neural behavior. Here, muscle torques are computed as:
\begin{eqnarray}
\tau_t = \tau_p (q, \dot{q}) + \tau_a (q, \dot{q}) \\
\tau_a (q, \dot{q}) = f_a(q) f_v(\dot{q}) \tau_{max} a
\end{eqnarray}
where, the torque at time $t$, $\tau_t$ is the summation of the torque due to passive musculotendon components, $\tau_p$, and active muscle contraction, $\tau_a$. Passive and active torques are functions of joint angle $q$ and joint rotational velocity $\dot{q}$. Muscle activation $a$ is computed from the excitation-activation dynamics as per \cite{Thelen2003}:
\begin{eqnarray}
\dot{a} &=& (e-a)\left[\frac{e}{tc_a} + \frac{1-e}{tc_d}\right], \mbox{ e $\geq$ a} \\
\dot{a} &=& \frac{e-a}{tc_d}, \mbox{ e $<$ a}
\end{eqnarray}
where, $e$ denotes the neural excitation, and $tc_a = 0.011$, $tc_d = 0.068$ the activation and deactivation time constants. We approximate the active-torque-angle and torque-velocity curves as 2D $5^{th}$ order Bezier curves, Fig. \ref{fig:humanModel}c. A smooth function behavior was an important prerequisite to the subsequent treatment in the optimal control problem.
\subsection{Rigid-Body Dynamics}
\label{subsec:rgbd}
Dynamics computation of the multi-body system is done using the open-source C++ library RBDL - Rigid Body Dynamics Library \cite{Felis2016}. RBDL builds on the concept of Spatial Algebra \cite{Featherstone2008} and provides efficient implementations of state-of-the-art multi-body dynamics algorithms.
\subsection{Parametrized Exoskeleton Models}
\label{sec:paramExo}
\begin{figure}[h]
\centering
\includegraphics[width=\linewidth]{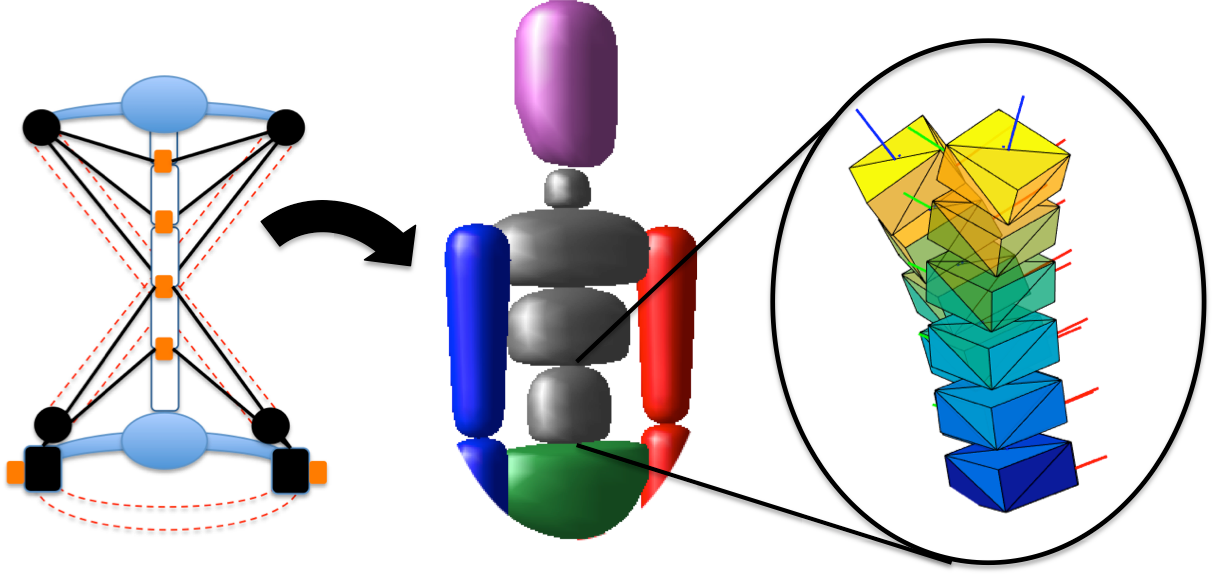}
 \caption{Ongoing work is focused on developing more complex models of spinal joints and exoskeletons.}
\label{fig:perspectives}
\end{figure}

Exoskeletons such as actuated lower-body devices or those to support spinal loads (Fig. \ref{fig:perspectives}) are parametrized with respect to size, weight, spring-damper dynamics and actuator capabilities. In addition to the dynamics of the exoskeleton, we also model the interaction between the human and the exoskeleton with rigid or compliant contacts. This parametrization follows from related recent work in the ORB research group, on design of lower limb exoskeletons using optimal control \cite{Koch2015}.
\begin{figure}[h]
\centering
\includegraphics[width=0.95\linewidth]{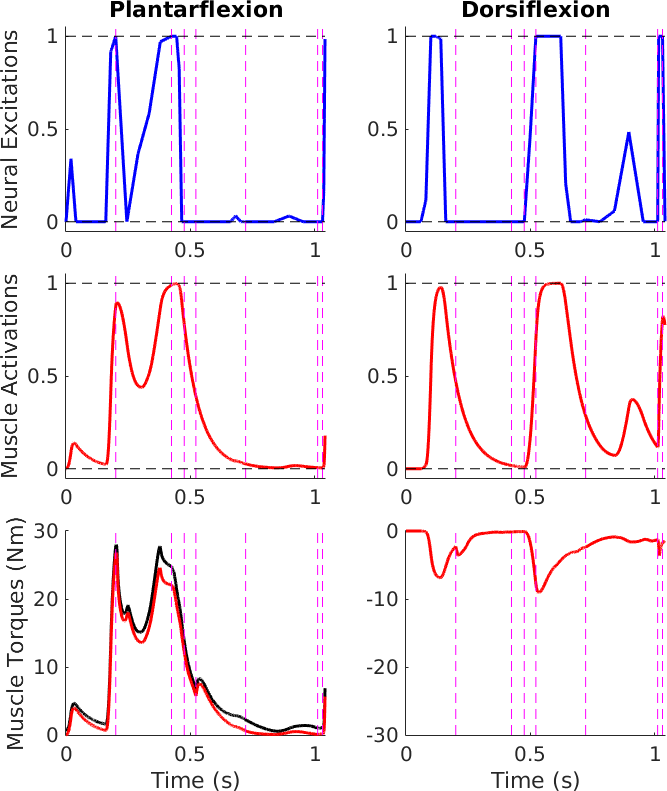}
 \caption{Neuromuscular dynamics at the ankle joint during walking. Top panels plot the neural excitations, middle panels the muscle activations, and bottom panels the muscle torques. Vertical lines indicate the change of model stages due to stepping during walking.}
\label{fig:resultsOCP}
\end{figure}
\section{Results}
\label{sec:results}
The optimal control problem (OCP) corresponding to model dynamics described above is solved using the direct multiple shooting method \cite{Bock1984} implemented in the software package MUSCOD-II \cite{Leineweber2003}. The OCP consists of several model stages with corresponding constraints based on the changing dynamics of the multi-body model and the interaction with the environment (e.g. foot contacts during stepping, hand contacts during lifting etc).

The controls to be identified correspond to the neural excitations, $e$. These controls were discretized as piecewise linear functions with additional continuity conditions between model stages. Fig. \ref{fig:resultsOCP} plots the excitations, activations and computed muscle torques for the agonist-antagonist pair at the ankle joint. Details of the OCP formulation corresponding to human movement models similar to the one developed here, can be found elsewhere \cite{Felis2016,Mombaur2009}.
\section{Discussion}
\label{sec:discussion}
The modeling and optimal control framework summarized here, provides a powerful basis to approximate the human body dynamics as well as to generate novel movements. Ongoing  work is focused on developing more complex models of joint kinematics (e.g. a complex spine) and exoskeletons, Fig. \ref{fig:perspectives}. We are also developing methods to identify subject-specific torque muscle parameters, as well as exoskeleton parameters by treating them as variables of the OCP. Another focus is to compute the  relative movements and contact forces between the exoskeleton and the user, in order to make the human-machine interface as comfortable as possible.

\end{document}